\documentclass[conference]{IEEEtran}
\IEEEoverridecommandlockouts
\usepackage{cite}
\usepackage{amsmath,amssymb,amsfonts}
\usepackage{algorithmic}
\usepackage{graphicx}
\usepackage{textcomp}
\usepackage{xcolor}
\def\BibTeX{{\rm B\kern-.05em{\sc i\kern-.025em b}\kern-.08em
    T\kern-.1667em\lower.7ex\hbox{E}\kern-.125emX}}

\usepackage{times}
\usepackage{soul}
\usepackage{url}
\usepackage[hidelinks]{hyperref}
\usepackage[utf8]{inputenc}
\usepackage[small]{caption}
\usepackage{graphicx}
\usepackage{amsmath}
\usepackage{amsthm}
\usepackage{booktabs}
\usepackage{algorithm}
\usepackage{algorithmic}
\usepackage[switch]{lineno}
\usepackage{cleveref}
\usepackage{amsmath}
\usepackage{amssymb}
\usepackage{subcaption}
\usepackage{paralist}
\usepackage[T1]{fontenc}

\def\R{\mathbb{R}}
\def\our{HyperTab}

\begin{document}

\title{HyperTab: Hypernetwork Approach for Deep Learning on Small Tabular Datasets}

\makeatletter
\newcommand{\linebreakand}{%
  \end{@IEEEauthorhalign}
  \hfill\mbox{}\par
  \mbox{}\hfill\begin{@IEEEauthorhalign}
}
\makeatother



\author{\IEEEauthorblockN{Witold Wydmański}
    \IEEEauthorblockA{\textit{Faculty of Mathematics and Computer Science} \\
    \textit{ Jagiellonian University}\\
    Kraków, Poland \\
    witold.wydmanski@uj.edu.pl}
    \and
    \IEEEauthorblockN{Oleksii Bulenok}
    \IEEEauthorblockA{\textit{Faculty of Mathematics and Computer Science} \\
    \textit{ Jagiellonian University}\\
    Kraków, Poland \\
    oleksii.bulenok@gmail.com}
    \linebreakand 
    \IEEEauthorblockN{Marek Śmieja}
    \IEEEauthorblockA{\textit{Faculty of Mathematics and Computer Science} \\
    \textit{ Jagiellonian University}\\
    Kraków, Poland \\
    marek.smieja@uj.edu.pl}
}

\maketitle

\begin{abstract}
    Deep learning has achieved impressive performance in many domains, such as computer vision and natural language processing, but its advantage over classical shallow methods on tabular datasets remains questionable. It is especially challenging to surpass the performance of tree-like ensembles, such as XGBoost or Random Forests, on small-sized datasets (less than 1k samples). To tackle this challenge, we introduce HyperTab, a hypernetwork-based approach to solving small sample problems on tabular datasets. By combining the advantages of Random Forests and neural networks, HyperTab generates an ensemble of neural networks, where each target model is specialized to process a specific lower-dimensional view of the data. Since each view plays the role of data augmentation, we virtually increase the number of training samples while keeping the number of trainable parameters unchanged, which prevents model overfitting. We evaluated HyperTab on more than 40 tabular datasets of a varying number of samples and domains of origin and compared its performance with shallow and deep learning models representing the current state-of-the-art. We show that HyperTab consistently outranks other methods on small data (with statistically significant differences) and scores comparable to them on larger datasets.
    
\end{abstract}


\section{Introduction}

Deep learning has already gained great success in various fields, such as computer vision \cite{krizhevsky_imagenet_2017}, natural language processing \cite{young_recent_2018}, and video analysis \cite{michelsanti_overview_2021} or reinforcement learning \cite{atari}. However, in tabular data analysis, deep learning methods are not as popular as in other areas. Somehow, although neural networks were first created with this aim in mind, it turned out that their performance on tabular data is subpar to other, much simpler algorithms \cite{shwartz2022tabular}, \cite{grinsztajn2022tree}.

There are many potential reasons for this. Modern deep learning architectures designed for computer vision, such as convolutional networks \cite{convnets} or vision transformers \cite{vision_transformer}, emerged after years of research to create inductive biases that match invariances and spatial dependencies of image data. Finding corresponding invariances in tabular data is hard, which makes the fully-connected architectures the first choice for tabular datasets. Moreover, typical computer vision models containing millions of parameters are trained on an enormous amount of data coming from common domain, such as photographs, which allows them to discover sophisticated patterns without overfitting. In real-world settings, small tabular datasets are ubiquitous. If the dimension of data is relatively large compared to the number of examples, then the fully-connected networks rapidly overfit, which prevents from using deeper architectures. 

In our work, we take inspiration from classical ensemble models, such as Random Forests \cite{breiman2001random} or random subspace method \cite{ho1998random}, which significantly improve the generalization ability of decision trees even for small high-dimensional data. Instead of training an individual model, Random Forests generate an ensemble of trees, each one taking into account a selected subset of features and specialized for a subset of samples. The feature subsetting plays the role of data augmentation, which generates multiple views of a single instance, allowing the number of available data to be increased.
Following the above reasoning, we introduce \our{}, a novel and effective technique for building an ensemble of neural networks for learning from small tabular datasets, see \Cref{fig:overview}. \our{} combines the ensembling strategy with the augmentation mechanism, which significantly increases the number of training data and, consequently, allows the use of larger network architectures. 
To meet the requirement that the augmentation is a class-invariant transformation (the class label does not change after applying the augmentation), we use the feature subsetting as the admissible augmentations. 

\our{} follows the hypernetwork approach, in which a single hypernetwork builds an ensemble of target networks. Given the augmentation identifier (subset of features), the hypernetwork generates the parameters of the target network, which is crafted to process data points transformed by this augmentation. In the case of feature subsetting augmentations, each target network operates on a lower-dimensional view, where data are represented by the subset of features. In this way, we have as many training examples as the number of $[augmentation, example]$ pairs. Moreover, since the parameters of target networks are not optimized but returned by the hypernetwork, we significantly reduce the number of trainable parameters compared to the size of the ensemble. The only trainable parameters are the weights of the hypernetwork. In consequence, we obtain an effective and modern framework for constructing a deep learning model for challenging cases of small tabular datasets. Finally, generating individual target networks for specific subsets of features using a single hypernetwork is also a way to automatically design network architectures, which include internal dependencies between coordinates.

To validate our approach, we used a diverse benchmark of 22 public tabular datasets and 20 real microbial data. Our experiments clearly confirm that \our{} gives a superior performance on small datasets -- its advantage over current state-of-the-art models being statistically significant. In the case of large datasets, where the problem of overfitting is reduced, \our{} works similarly to other algorithms. In addition to benchmark tasks, we performed a detailed analysis of \our{}, which allows us to understand how the selection of hyperparameters affects its performance.

Our contributions can be summarized as follows:
\begin{itemize}
\item We introduce \our{}, which effectively builds an ensemble of neural networks using the hypernetwork approach.
\item We apply an augmentation strategy based on feature subsetting, which is consistent with the characteristic of tabular data, and allows for virtually increasing the number of training samples.
\item We evaluate \our{} on various datasets and show that it obtains the state-of-the-art on small data sets.
\end{itemize}
We make a Python package with the code available to download at \url{https://pypi.org/project/hypertab/}.

\begin{figure}
    \centering
        \includegraphics[width=\linewidth]{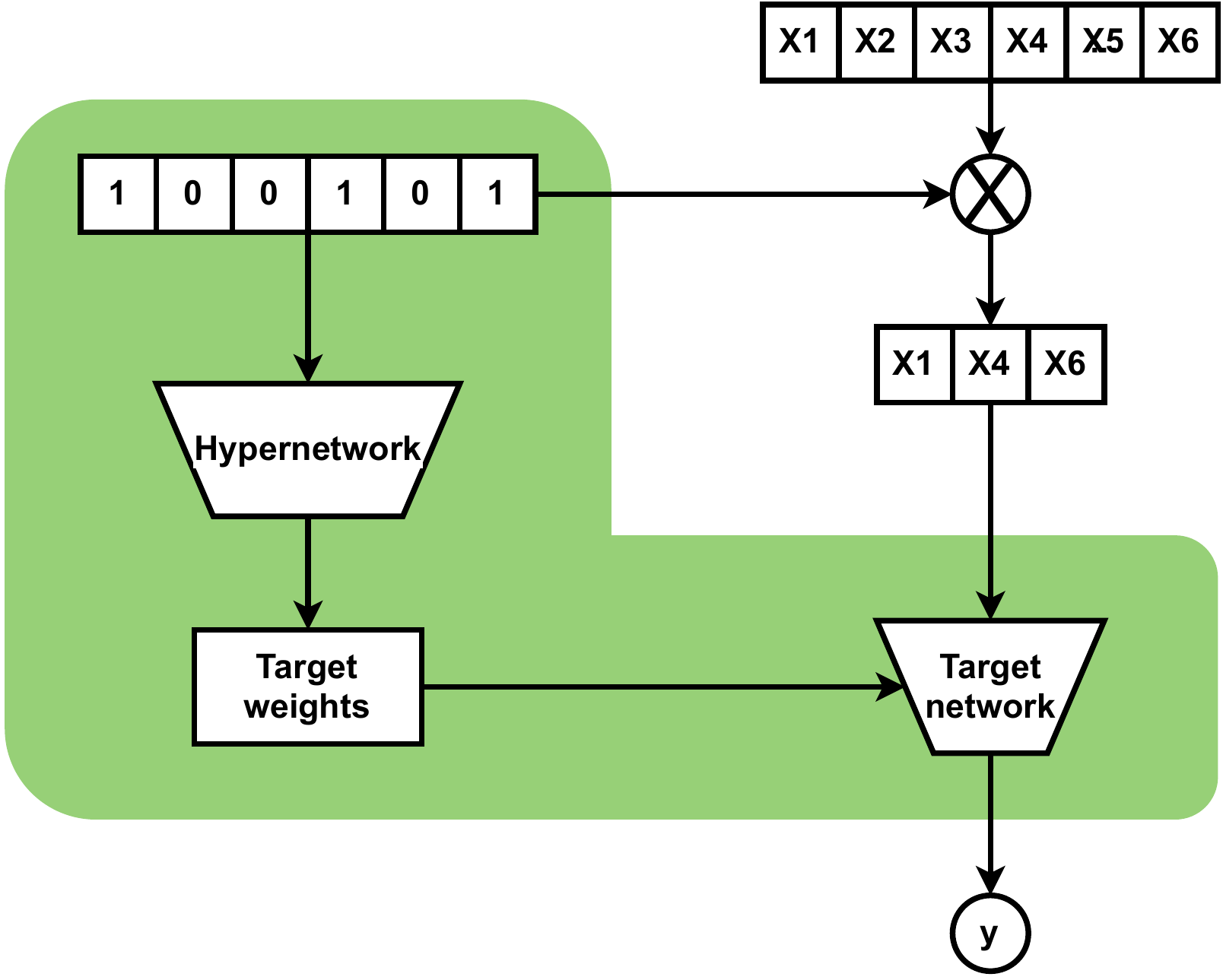}
    \caption{{\bf A general \our{} architecture.} Given a binary mask representing a subset of features (augmentation), the hypernetwork generates the weights to the target network, which operates on a lower-dimensional view of data determined by the mask. The response of \our{} is based on multiple target networks generated for individual augmentations.}
    \label{fig:overview}
\end{figure}

\section{Related work}

\subsection{Hypernetworks}

Hypernetwork is hardly a new concept -- having been introduced in \cite{ha_hypernetworks_2016} they have already been successfully used in various domains. In general, hypernetworks are an effective alternative to conditional neural networks \cite{galanti2020modularity} -- the condition represents the input to the hypernetwork, which generates the weights of the target networks responsible for solving a target task. Hypernetworks appear frequently in the meta-learning literature \cite{zhao2020meta}. In \cite{von2019continual}, hypernetworks were used to tackle the problem of \textit{catastrophic forgetting} -- an observation that networks that are trained sequentially on multiple tasks tend to vastly underperform in comparison to task-specific training. The authors of \cite{10.1007/978-3-030-30493-5_48} and \cite{skorokhodov2021adversarial} use hypernetworks to construct a functional representation of images, in which we can inspect the image at various resolutions and perform on it arbitrarily continuous operations. \cite{pszemek} shows that hypernetworks can also be used to create generative models for 3D point clouds. 

\subsection{Augmentations}

Data augmentation is a key component of current deep learning models, which allows learning meaningful representation even in a self- or semi-supervised setting. Although there is a standard set of augmentations for computer vision or natural language \cite{nl-augmenter}, there is no consensus on how to select proper augmentations for tabular datasets, in which spatial or semantic structures usually do not exist. One of the standard approaches relies on corrupting data, e.g. by adding Gaussian noise \cite{vincent2008extracting}. Alternatively, we can replace the values of selected features (determined by the mask vector) with the dummy values \cite{ucar2021subtab}. We can either zero out them, perform the imputation using the mean or median, as well as use values taken from other instances. Although analogous strategies have been applied in self-supervised learning or in the definition of pretext tasks \cite{fakoor2020fast}, \cite{yoon2020vime} \cite{ucar2021subtab}, they do not guarantee that the class label remains unchanged after such augmentations. The authors of \cite{darabi2021contrastive} applied mix-up training by mapping samples to a low-dimensional latent space and encouraging interpolated samples to have high similarity
within the same labeled class. In our paper, we decided to construct a lower-dimensional view of the original data by randomly selecting a subset of features, which is analogous to the idea of the random subspace method. However, instead of training many individual models, we employ the hypernetwork, which is responsible for generating weights to target models. In this way, we reduce the number of trainable parameters. 

\subsection{Shallow models for tabular datasets}

In contrast to computer vision or natural language processing, shallow models, such as Support Vector Machines \cite{cortes1995support}, \cite{scholkopf2002learning}, Random Forests \cite{ho1995random}, \cite{breiman2001random}, and Gradient Boosting \cite{friedman2001greedy}, are usually the first choice for learning from tabular datasets. In particular, the family of Gradient Boosting algorithms \cite{friedman2001greedy}, including XGBoost \cite{chen2016xgboost}, LightGBM \cite{ke2017lightgbm}, CatBoost \cite{prokhorenkova2018catboost}, and GOSDT \cite{DBLP:journals/corr/abs-2006-08690} achieve impressive performance and frequently exceed the performance of deep learning models. Both Gradient Boosting as well as Random Forests generate an ensemble of weak learners composed of decision trees, but they differ in the way those trees are built and combined.

\subsection{Deep learning on tabular data}

To take advantage of the flexibility of neural networks, various architectures have recently been proposed to improve their performance on tabular data. Inspired by CatBoost, NODE performs a gradient boosting of oblivious decision trees, which is trained end-to-end using gradient-based optimization \cite{node19}. The aim of Net-DNF is to introduce an inductive bias in neural networks corresponding to logical Boolean formulas in disjunctive normal forms \cite{katzir2020net}. It encourages localized decisions, which involve small subsets of features. TabNet uses a sequential attention mechanism to select a subset of features, which are used at each decision step \cite{arik2021tabnet}. Hopular is a deep learning architecture in which every layer is composed of continuous modern Hopfield networks \cite{schafl2022hopular}. The Hopfield modules allow one to detect various types of dependencies (feature, sample, and target) and have been claimed to outperform concurrent methods of small and medium-sized datasets. The authors of \cite{kadra2021well} show that the key to boosting the performance of deep learning models is the application of various regularization techniques. They show that fully connected networks can outperform concurrent techniques by applying an extensive search on possible regularizers. Although authors of recent deep learning models often claim to outperform shallow ensemble models, other experimental studies seem to deny these conclusions, showing that XGBoost with careful hyperparameter tuning presents superior performance \cite{shwartz2022tabular}, \cite{grinsztajn2022tree}. 

In our paper, we are especially interested in overcoming the limitations of deep learning models on small tabular datasets. Although learning from tabular data also poses other challenges, such as encoding categorical features, we intentionally focus on that specific case. By taking advantage of feature subsetting used in decision tree ensembles and hypernetworks, we build a modern deep learning model which is especially suited for small sample problems.

\section{The \our{} model}

It has been widely noted \cite{grinsztajn2022tree} that tree-like ensemble methods often outperform deep learning algorithms on tabular datasets. Following this observation, we introduce \our{}, which follows these algorithms and combines an implicit data augmentation with neural network ensembles to ensure its performance on small datasets. 

\subsection{Model overview}

\our{} consists of two main components: hypernetwork $H$ and an ensemble of target networks $T_j$, for $j=1,\ldots,k$, see \Cref{fig:overview}. The hypernetwork takes the type of augmentation as input and returns the parameters of the target network, which is designed to use such an augmented view of data. Since augmentations are defined as feature subsetting, every target network operates on a lower-dimensional view of the data determined by selected coordinates. In contrast to typical neural network ensembles, the weights of target networks are not optimized directly using gradient descent but are generated by the hypernetwork. The only trainable parameters are the hypernetwork weights. 

Such an approach is especially profitable for small datasets, because augmentations allow us to significantly increase the number of training examples. Instead of using raw data, we combine each data point with all types of augmentations, giving us $n \cdot k$ training examples, where $n$ is the number of examples, and $k$ denotes the number of augmentations. The number of trainable weights of \our{} remains roughly the same as in typical fully connected neural networks processing original data. 

In the following parts, we describe our approach in detail.

\subsection{Augmentations}

It is not obvious how to construct augmentations suitable for tabular datasets. Augmentation is a class-invariant transformation of the data, which means that the class label cannot change after applying this transformation. Although recent progress in computer vision delivered a great variety of augmentations, there is no gold standard for tabular data. For this reason, we decided to restrict our attention to feature subsetting as admissible augmentations. By representing a data point by the subset of its features, we do not introduce noise, but only limit the information contained in the original data. 
Using an analogy with computer vision, feature subsetting is similar to random cropping. In our case, however, we reduce the dimension of the original data and completely eliminate features that have not been selected. An in-depth analysis of the relationship between feature subsetting and performance of the algorithm has been performed in \cref{sec:an}.

Let $X=\{x_1, x_2, \dots, x_n\} \subset \R^d$ be a tabular dataset, which we want to use in training a deep learning model. By $c \subset \{1,\ldots,d\}$, where $|c| = l$, we denote the subset of $l$ selected indices. Applying the augmentation defined by $c$ to a sample $x \in \R^d$ produces a vector $x[c] \in \R^l$, which represents a lower-dimensional view of the data point $x$. 

\subsection{Construction of the ensemble}

Every target network (component model in the \our{} ensemble) is designed to process a specific augmented view of the data. More precisely, the target network $T_c: \R^l \to Z$ takes a lower-dimensional representation $x[c]$ and returns a vector $z \in Z$, e.g. logits in the case of classification. The vector $z$ can be converted to the final target value, e.g. a class label $y \in Y$. The augmentation $c$ determines the form of the target network $T_c$. 

Instead of training an individual target network $T_c$ using gradient descent, we construct a central mechanism to generate the whole ensemble. That is, we use a hypernetwork $H$, which returns the parameters of the target network for a given type of augmentation. Since we work with feature subsetting, the augmentation $c \subset \{1,\ldots,d\}$ can be encoded as a binary mask $m \in \{0,1\}^d$ indicating the selected features, i.e. 
\begin{equation}\label{eq:mask}
m_j = \left\{
\begin{array}{ll}
     1, & \text{ for } j \in c,  \\
     0, & \text{ otherwise.}
\end{array}
\right.
\end{equation}
The hypernetwork is thus a neural network ${H_\psi: \{0,1\}^d \to \Theta}$, which transforms a binary mask $m$ representing the augmentation $c$ to the weights $\theta_c$ of the target network $T_{\theta_c}$, i.e.
$$
\theta_c = H_\psi(m).
$$
The architecture is common for all target networks, but their weights are individually generated by the hypernetwork to process a specific augmented view of the data. Every target network returns: 
$$
z_c = T_{\theta_c}(x[c]),
$$
given an augmented view of $x \in \R^d$.

\subsection{Training}

To train \our{}, we optimize the weights $\psi$ of the hypernetwork $H_\psi$. As mentioned, we do not optimize the parameters of the target networks directly, but only the weights of hypernetwork. 

In a training step, we take a minibatch of augmentations $\mathcal{B}_c = \{c_1,\ldots,c_a\}$, where $c_j \subset \{1,\ldots,d\}$ such that ${|c_j| = l < d}$, and define the corresponding masks ${m_j \in \{0,1\}^d}$ as in \eqref{eq:mask}. Using the hypernetwork, we generate the weights of the target networks $\theta_j = H_\psi(m_j)$. Every target network is then applied to the minibatch of data points $\mathcal{B}_x = \{x_1,\ldots,x_b\}$ producing partial predictions $z_{ij} = T_{\theta_j}(x_i[c_j])$, for $i=1,\ldots,b$ and $j=1,\ldots,a$. Vectors $z_{ij}$ are compared to true targets $y_{i}$ via a given loss function $\mathcal{L}(y_i,z_{ij})$, e.g., cross-entropy with softmax in the case of classification. The loss is minimized by changing the parameters $\psi$ of the hypernetwork $H_\psi$ using gradient descent.

As can be seen, a training sample is a pair of augmentation and data point. As a consequence, we have as many training samples as the number of data points times the number of augmentations. This is especially useful for small datasets because we can significantly increase the number of training data. Since the number of trainable network weights is comparable to that of a typical neural network, our approach prevents the model from overfitting.

\subsection{Inference}

Once trained, the hypernetwork $H$ can produce the ensemble of weak learners $\theta_j = H_\psi(m_j)$,
where $m_j$ is a mask corresponding to the augmentation $c_j$, for $j=1,\ldots,a$. The final prediction for a given sample $x \in \R^d$ is calculated as the average of the predictions of the target networks taken over all augmentations:
$$
z = \frac{1}{a} \sum_{j=1}^a T_{\theta_j}(x[c_j]).
$$
We emphasize that $z_c = T(x[c])$ is the result of the last layer of the target network, for example, logits in the case of classification. We can also use different aggregation methods, but the mean pooling applied to logits makes \our{} robust to the noisy augmentations containing irrelevant features, as shown in \Cref{sec:an}.

\section{Experiments}

\subsection{Benchmark}

\paragraph{Experimental setup}

In order to thoroughly check the potential of \our{}, we check its performance on multiple datasets coming from diverse domains. For transparency, we consider classification problems, but \our{} can also be applied to regression tasks. We distinguish small datasets, in which the number of samples is less than 1k, and larger datasets with more than 1k samples. Since the construction of \our{} is suitable for a small sample problem, we expect it to reach the state-of-the-art at least in the first group of datasets. An overview of the datasets can be found in Table 1 of the Supplementary Materials.

\begin{table*}[t]\centering
\caption{Performance of the algorithms on: (top) small datasets ($n \leq 1k$); (bottom) medium and large datasets ($n > 1k$). We report the average of 5 runs and the standard deviation in brackets.}
    \label{table:results1}
    \setlength{\tabcolsep}{12pt} 
    \footnotesize
    \begin{tabular} {@{}l}
  $\left.
  \begin{tabular}{p{3.5cm}p{1.7cm}p{1.7cm}p{1.7cm}p{1.7cm}p{1.7cm}}
    \toprule
    {Dataset} & {XGBoost} & {DN} & {RF} & {HyperTab}  & {Node} \\
    \midrule
     {Breast Cancer}  & 93.85 (1.44)  & 95.58 (1.04)  & 95.96 (1.52)  & \bfseries 97.58 (1.11) & 96.19 (1.11) \\
     {Connectionist}    & 83.52 (3.94)  & 79.02 (5.29)  & 83.50 (5.55)  & \bfseries 87.09 (5.53) & 85.61 (3.48) \\
     {Dermatology}  & 96.05 (0.89)  & 97.80 (1.17)  & 97.21 (1.66)  & 97.82 (1.24) & \bfseries 97.99 (1.20) \\
     {Glass} & 94.74 (3.91)  & 46.96 (2.56)  & 97.02 (1.51)  & \bfseries 98.36 (3.21)  & 44.90 (1.90) \\
     {Promoter} & 81.88 (5.59)   & 78.91 (3.93)  & 85.94 (6.79)  & \bfseries 89.06 (5.41) & 83.75 (4.64) \\
     {Ionosphere} & 90.67 (2.75) & 93.43 (3.72) & 92.43 (2.60) & \bfseries 94.52 (1.47) & 91.03 (1.79) \\
     {Libras} & 74.38 (4.55) & 81.54 (3.99) & 77.42 (3.88) & \bfseries 85.22 (2.92) & 82.72 (3.27) \\
     {Lymphography} & 85.94 (3.14) & 85.74 (5.28) & \bfseries 87.19 (4.33) & 83.90 (5.01) & 83.93 (5.82) \\
     {Parkinsons} & 86.35 (4.77)     & 74.96 (4.90)  & 86.84 (6.26) & \bfseries 95.27 (3.06) & 80.20 (5.29) \\
     {Zoo}       & 92.86 (8.75)  & 72.62 (4.96)  & 92.62 (7.97)  & \bfseries 95.27 (3.06)  & 89.05 (3.98)    \\
     {Hill-Valley without noise}       & 65.53 (0.00)  & 56.39 (2.89)  & 57.33 (0.00)  & \bfseries 70.59 (4.90) & 52.71 (0.34)    \\
     {Hill-Valley with noise}       & 58.45 (0.00)  & 56.06 (1.65)  & 55.66 (0.00)  & \bfseries 70.16 (3.25)  & 51.09 (0.26)    \\
     {OvarianTumour} & 60.61 (8.80)   & 33.33 (0.00) & 51.24 (7.53)    & \bfseries 76.60 (4.48)  & 68.39 (10.82) \\
     {Heart Disease (Cleveland)} & 79.17 (7.24) & 82.62 (4.50) & 81.10 (3.89) & \bfseries 83.33 (2.54) & 82.38 (4.59) \\
     \midrule
    {Mean rank} & 3.50 & 3.78 & 3.07 & 1.35 & 3.29 \\
    \midrule
    \end{tabular}
  \right\}\text{\rotatebox[origin=c]{270}{small datasets}}$\\[\bigskipamount]
     
      $\left.
  \begin{tabular}{p{3.5cm}p{1.7cm}p{1.7cm}p{1.7cm}p{1.7cm}p{1.7cm}}
     {FashionMNIST} & 89.45 (0.18) & 89.01 (0.04) & 88.04 (0.21) & \bfseries 90.22 (0.32) & 89.51 (0.21) \\
     {CNAE-9}       & 90.49 (2.05)  & \bfseries 94.97 (0.77)  & 91.85 (1.36)  &  92.25 (2.55)  & 94.72 (1.17)    \\
     {Multiple Features}   & 98.03 (0.44)  & 98.27 (0.61)  & \bfseries 98.98 (0.36)  &  98.12 (0.81)  & 98.58 (0.45)    \\
     {Devanagari} & 72.03 (0.58) & 75.24 (0.47) & 71.15 (0.73) & \bfseries 78.92 (0.66) & 78.20 (1.08) \\
     {Volkert} & \bfseries 63.48 (0.37) & 54.32 (1.51) & 58.08 (0.26) & 57.41 (2.38) & 59.25 (0.99) \\
     {Nomao} & \bfseries 96.50 (0.15) & 95.71 (0.30) & 95.83 (0.29) & 95.53 (0.27) & 95.23 (0.26) \\
     {Fabert} & 30.82 (0.78) & 27.09 (0.26) & 66.74 (1.00) & 60.09 (0.09) & \bfseries 67.47 (1.19) \\
     {Christine} & \bfseries 72.89 (0.98) & 71.80 (0.53) & 72.21 (0.98) & 72.42 (2.19)  & 71.78 (1.24) \\
    \midrule
    {Mean rank} & 3.00 & 3.50 & 3.12 & 2.75 & 2.62 \\
    \bottomrule
    \end{tabular}
  \right\}\text{\rotatebox[origin=c]{270}{larger datasets}}$
    \end{tabular}
\end{table*}

As an evaluation measure, we use balanced accuracy, which is especially designed for datasets with unbalanced classes. If the number of examples in each class is comparable, then balanced accuracy gives analogical scores to accuracy. In consequence, balanced accuracy is a perfect measure for our case, where datasets have diverse characteristics of classes.

The target networks in \our{} are fully connected networks with a single hidden layer, with 5 to 50 neurons (defined by a hyperparameter). The hypernetwork is made up of 3 hidden layers with 128, 64, and 64 neurons. All activations are defined by the ReLU function. The number of augmentations and the number of selected features are hyperparameters chosen in a grid search procedure.

We test the performance of \our{} along with two shallow algorithms and two deep learning models:
\begin{itemize}
    \item {\bf RF} (Random Forests). It is an ensemble of decision trees, which combines bagging with feature subsetting. 

    \item {\bf XGBoost} (Extreme Gradient Boosting). It is an implementation of the Gradient Boosting algorithm, which obtains the best performance among shallow machine learning models on tabular datasets. 

    \item {\bf DN} (Fully connected neural network with dropout regularization). We use a typical fully connected architecture with dropout regularization to avoid overfitting on small datasets. Dropout regularization before the first hidden layer makes the model similar to \our{}, in which an individual model is trained on a subset of features.
    
    \item {\bf NODE} (Neural Oblivious Decision Ensembles). It is a recent deep learning ensemble designed for tabular data, in which an individual decision tree is trained on a subset of features selected in a differentiable way. Comparing \our{} with NODE allows one to verify different subsetting mechanisms in both methods and the way of generating the ensembles.
\end{itemize}

All algorithms are subject to the optimization of hyperparameters separately for each of the datasets. Grids specific to the algorithms can be found in the Appendix. 
To reduce random effects, we report mean and standard deviation across 5 runs of the algorithms. In particular, \our{} is evaluated on 5 randomly selected sets of augmentations.

\paragraph{Results}

The results presented in Table \ref{table:results1} (top) clearly show that \our{} outperforms comparative methods on most examples of small datasets. In many cases, the difference between \our{} and the second best-performing method is extremely high (e.g., in Parkinson, Hill-Valley, and Ovarian Tumor datasets), which confirms the advantages of \our{} on small sample problems. There are two cases, where \our{} does not obtain the best results: on Dermatology \our{} is slightly worse than NODE, while on Lymphography the difference is greater.

To summarize the results, we apply statistical tests, see \cite{demvsar2006statistical}, specifically we used the Friedman test with Nemenyi post hoc analysis. For this purpose, we ranked all methods on every data set, i.e. the
best-performing method got ranked 1, the second-best method got ranked 2, etc. Given a ranking of the methods, the analysis consists of two steps:
\begin{compactitem}
    \item The null hypothesis is made that all methods perform the same and the observed differences are merely random (the hypothesis is tested by the Friedman test, which follows a $\chi^2$ distribution,
    \item Having rejected the null hypothesis, the differences in ranks are analyzed by the Nemenyi test.
\end{compactitem}

    

\begin{figure}[h]
    \centering
    \includegraphics[width=0.9\linewidth]{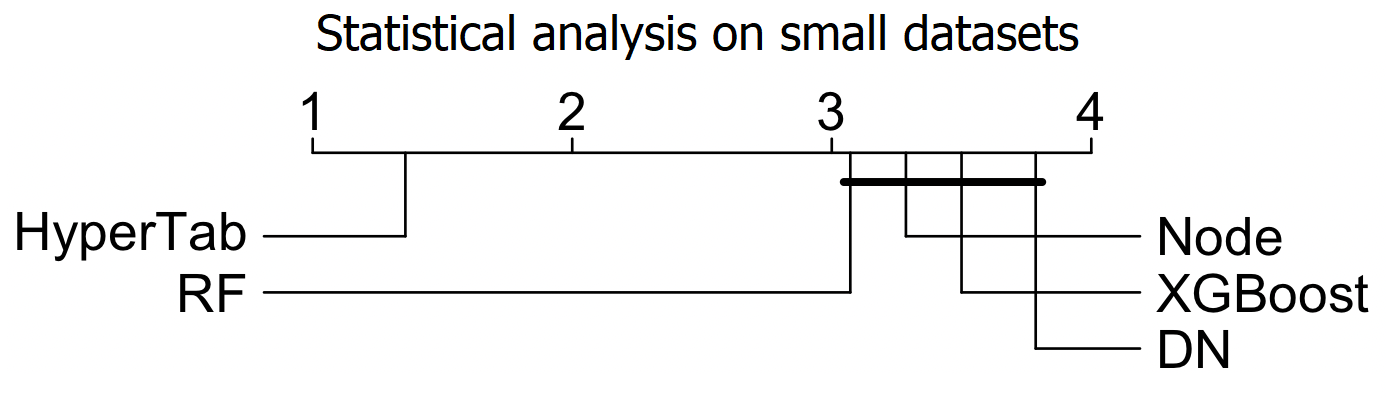}
    \includegraphics[width=0.9\linewidth]{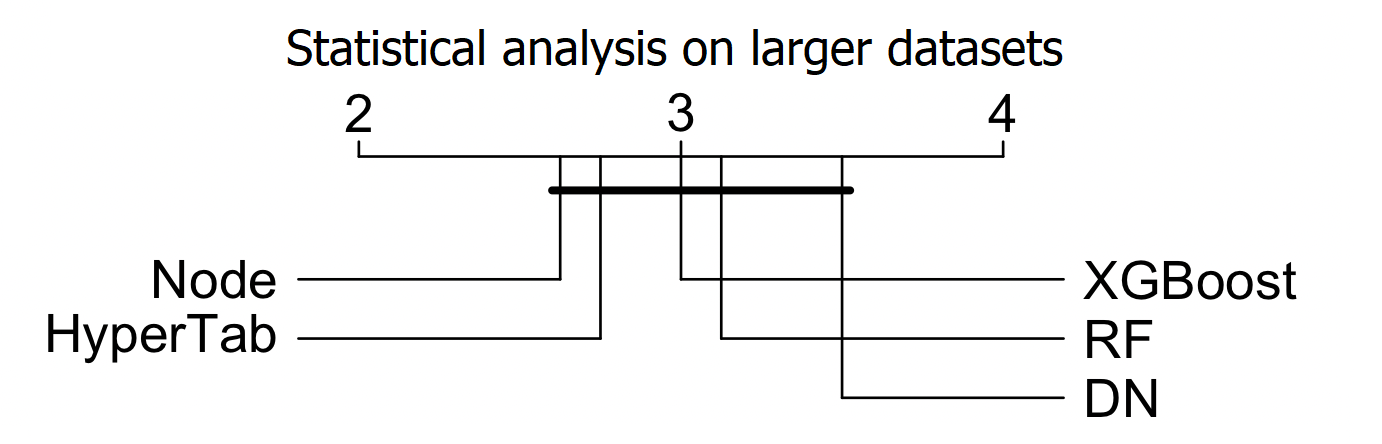}
    \caption{Statistical comparison of the methods on (top image) small datasets, (bottom image) lager datasets. A horizontal line connecting ranks shows which difference is not significant. \our{} is confirmed to perform statistically better than comparative methods on small datasets.}
    \label{fig:ranking_small}
\end{figure}

Figure \ref{fig:ranking_small} (top) visualizes the results for a significance level of $p = 0.05$. The x-axis
shows the mean rank for each method. The groups of methods for which the difference in mean rank is not statistically significant are connected by horizontal bars. As can be observed, the difference between \our{} and the second-best method is large. Moreover, the advantage of \our{} over all algorithms is statistically significant.

For completeness, we also perform the evaluation on larger datasets. As can be seen in Table \ref{table:results1} (bottom), \our{} performs on par with other methods, which confirms our initial hypothesis that \our{} is best suited to small datasets. It obtains the highest results on two datasets. However, it is difficult to indicate the best algorithm across all datasets. 
The statistical test shows that the differences between algorithms are not significant, see Figure \ref{fig:ranking_small} (bottom).


\begin{figure}[h]
    \centering
    \includegraphics[width=0.9\linewidth]{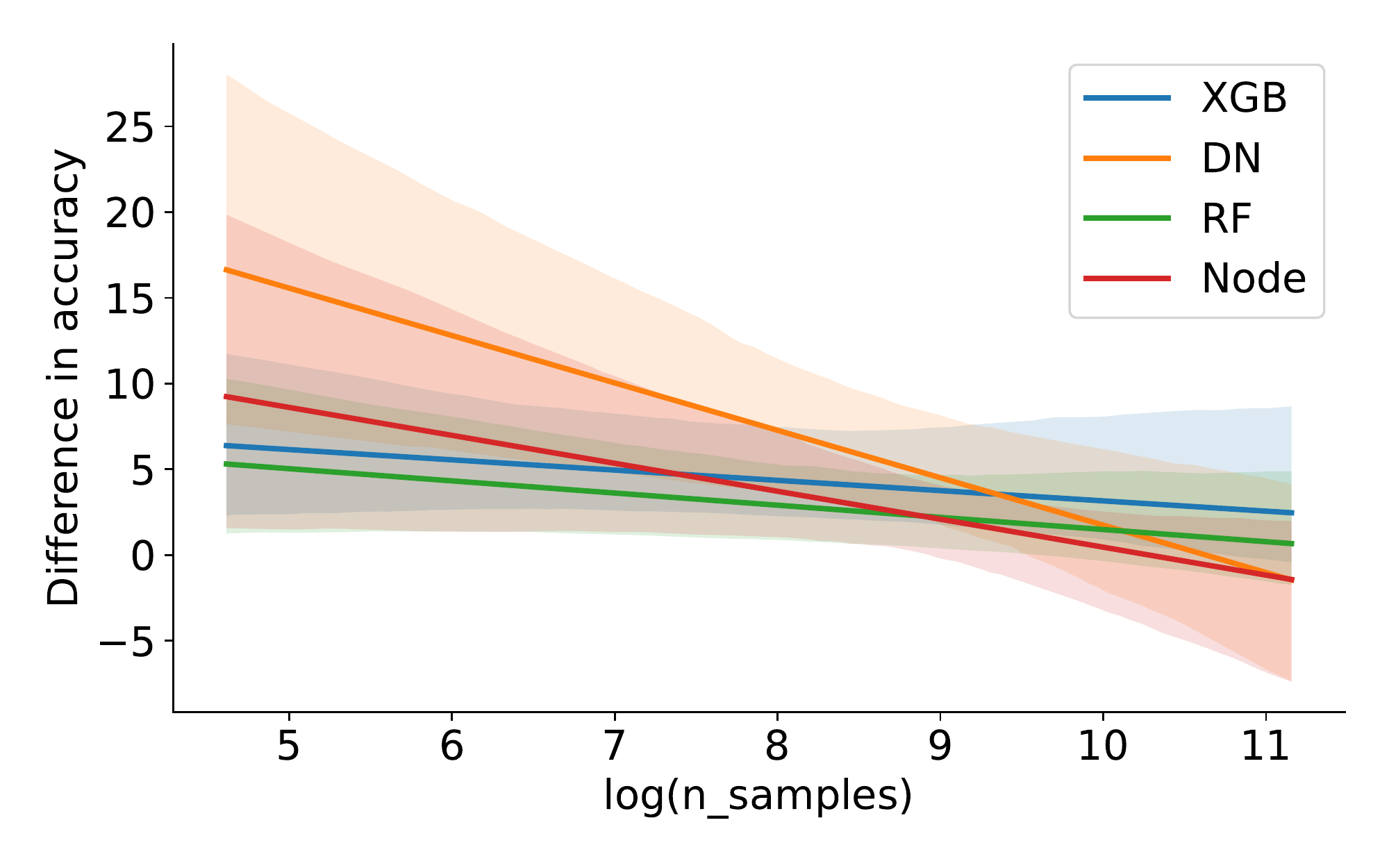}
    \caption{Difference in balanced accuracy between \our{} and other methods in function of the number of samples. The advantage of \our{} over all algorithms gradually increases as the number of samples decreases.}
    \label{fig:hypertab_diff2}
\end{figure}

\Cref{fig:hypertab_diff2} shows the relationship between the size of the dataset and the advantage of \our{} over other methods. Specifically, we report the difference in performance between \our{} and other methods in relation to the number of training samples. For transparency, we illustrate the estimated correlation. As can be seen, the advantage of \our{} over all methods is greater for small samples and gradually decreases for larger datasets. We verified that the estimated correlation factor calculated for RF, the second-best performing method, with respect to the mean rank, 
is statistically significant with $p$-value equal to 0.05. It further supports our hypothesis that \our{} is well suited for the classification of small tabular datasets. For large datasets, \our{} is outperformed by NODE, which is one of the SOTA deep learning models for tabular data.


\subsection{Use-case on microbial data}
To further validate the performance of \our{}, we tested it in a real-world scenario of metagenomic analysis. The purpose of this experiment is to assess what results we may expect to see once \our{} is adopted by researchers from domains, who do not necessarily possess machine learning expertise.

Following the procedure described in \cite{qu_application_2019}, we start from an OTU table: operational taxonomical units of microorganisms present in the samples. We then apply a rudimentary feature selection method by discarding constant features and preprocess the data using ANCOMBC \cite{lin_analysis_2020}. Since the procedure does not include tuning of the hyperparameters, we decided to also omit this step to replicate the real-world scenario as closely as possible.

We test \our{} against three other algorithms commonly used in microbiome classification \cite{qu_application_2019}: Random Forest, XGBoost, and Support Vector Classifier \cite{svc}. Our data consists of 20 datasets (summarized in the Supplementary Materials) that contain samples of the gut microbiome of patients with and without colorectal cancer. Their aim is to predict whether the patient has or does not have the aforementioned cancer based only on the composition of his gut microbiome.

\Cref{table:bio_datasets} contains results for each of the datasets, and \Cref{fig:ancombc} shows an overview of the scores. Once again, the results show that \our{} consistently outperforms other methods on small datasets. The difference between \our{} and the other methods is statistically significant.


\begin{table}[ht]
    \centering
    \caption{Results on the metagenomic datasets. Each dataset's name is derived from the first author's surname and the year of publication.}
    \label{table:bio_datasets}
    \setlength{\tabcolsep}{6pt}
    \footnotesize
    \begin{tabular}{lcccc}
    \toprule
    Dataset &    RF &   SVC &  XGBoost &  \our{} \\
    \midrule
    FengQ\_2015       &  0.84 &  0.87 &     0.81 &     \bfseries 0.98 \\
    GuptaA\_2019      &  0.80 &  0.90 &     0.82 &      \bfseries 0.98 \\
    HanniganGD\_2017  &  0.59 &  0.53 &     0.60 &      \bfseries 0.71 \\
    JieZ\_2017        &  0.73 &  0.79 &     0.79 &      \bfseries 0.93 \\
    KeohaneDM\_2020   &  \bfseries 0.60 &  0.44 &     0.49 &      0.50 \\
    LiJ\_2017         &  0.49 &  0.50 &     0.47 &      \bfseries 0.58 \\
    NagySzakalD\_2017 &  0.73 &  0.74 &     0.76 &      \bfseries 0.79 \\
    NielsenHB\_2014   &  0.72 &  0.71 &     0.73 &      \bfseries 0.78 \\
    QinJ\_2012        &  0.66 &  0.70 &     0.67 &      \bfseries 0.73 \\
    QinN\_2014        &  0.89 &  0.90 &     0.87 &      \bfseries 0.91 \\
    RubelMA\_2020     &  \bfseries 0.83 &  0.82 &     0.77 &      0.80 \\
    ThomasAM\_2018a   &  0.62 &  0.54 &     \bfseries 0.71 &      0.66 \\
    ThomasAM\_2018b   &  0.57 &  0.58 &     0.58 &      \bfseries 0.77 \\
    ThomasAM\_2019c   &  0.70 &  0.67 &     0.78 &      \bfseries 0.88 \\
    VogtmannE\_2016   &  \bfseries 0.72 &  0.56 &     0.70 &      0.64 \\
    WirbelJ\_2018     &  0.72 &  0.85 &     0.76 &      \bfseries 0.92 \\
    YachidaS\_2019    &  0.65 &  0.64 &     0.65 &      \bfseries 0.79 \\
    YuJ\_2015         &  0.68 &  0.75 &     0.73 &      \bfseries 0.77 \\
    ZellerG\_2014     &  0.68 &  \bfseries 0.75 &     0.68 &      0.66 \\
    ZhuF\_2020        &  0.72 &  \bfseries 0.79 &     0.72 &      0.70 \\
    \midrule
    Mean rank & 2.97 & 2.7 & 2.73 & 1.6 \\
    \bottomrule
    \end{tabular}
\end{table}

\begin{figure}
    \centering
    \includegraphics[width=0.8\linewidth]{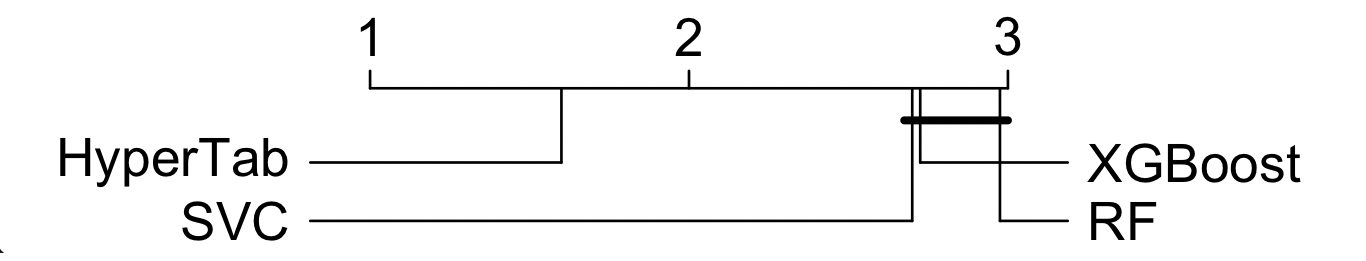}
    \caption{Analysis performed on microbial datasets shows that the difference between \our{} and other algorithms is statistically significant.}
    \label{fig:ancombc}
\end{figure}

\subsection{Analysis} \label{sec:an}

\paragraph{Dependence on the selection of augmentations}

There are two main parameters that influence the performance of \our{}: the number of augmentations (target models) and the number of features selected by each augmentation. In this experiment, we analyze their optimal values on two versions of the F-MNIST datasets. 

First, we consider F-MNIST with only 100 samples (10 per class), which corresponds to the small sample problem. As can be seen in \Cref{fig:mnist} (top), the highest accuracy is obtained for a relatively large number of augmentations (80-200) and a small number of selected features (20-50). The large number of augmentations virtually increases the amount of training data, while the small number of selected features allows for constructing an ensemble with diverse target models.


To extend our analysis, we performed an analogical experiment on the full F-MNIST dataset containing 60k training samples. In contrast to the previous case, here we observe that the optimal performance is obtained for a smaller number of augmentations and a higher number of selected features, see \Cref{fig:mnist} (bottom). It may follow from the fact that for large datasets, we do not need to virtually increase the number of training samples as in the case of small sample problems. However, if the number of target networks is small, we need to use many features in each augmentation to provide enough information about the data.

In conclusion, the number of augmentations should be selected jointly with the number of features in each augmentation. For small-sample problems, it is beneficial to use a high number of augmentations with a small number of features, while for larger datasets this relation should be inverse. 



\begin{figure}[htb]
    \centering
    \includegraphics[width=0.45\textwidth]{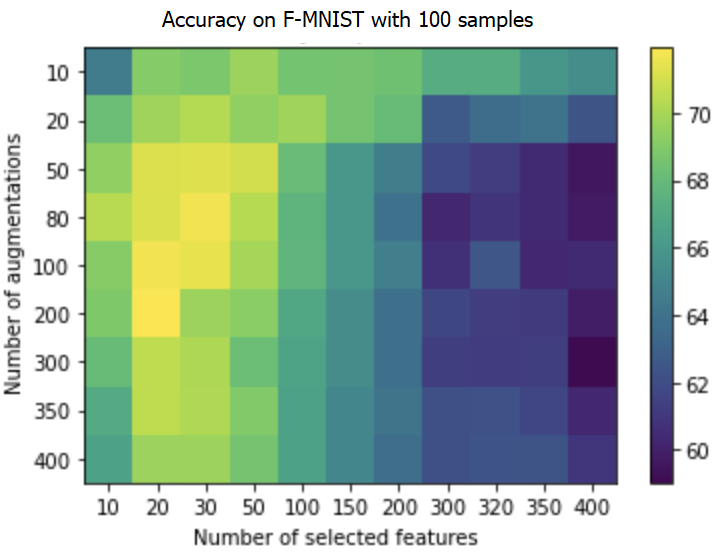}
    \includegraphics[width=0.45\textwidth]{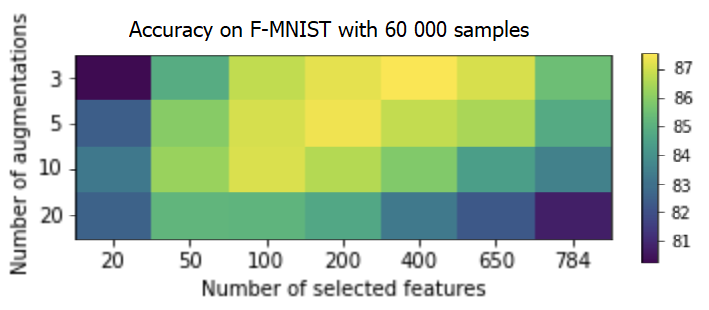}
    \caption{Influence of the number of augmentations and the number of selected features on the \our{} performance analyzed on two versions of F-MNIST. Smaller datasets strongly benefit from the proposed ensembling strategy.}
    \label{fig:mnist}
\end{figure}

\paragraph{Dependence on irrelevant features}

A natural question arises for \our{}: What if there are few important features and many augmentations, by chance, happen to omit them? F-MNIST allows testing that phenomenon to some degree, since many of their pixels are purely background noise, and there is a high chance that for small number of selected features most of them will consist of noise only. 

To further investigate this scenario, we devised a synthetic dataset with 50 columns, 49 of them randomly sampled from uniform distribution, and one being linearly dependent on the class of the example. The dataset has 5 unique classes and 50 samples. Each augmentation selects 10 features. We verified that only 17\% of the augmentations use the informative feature, while the rest of them rely on noise for their predictions. Despite that, \our{} was able to achieve perfect accuracy on this dataset. 

To further analyze the reason behind this behavior, we inspect the predictions of individual target networks and present their histograms in \Cref{fig:ablation}. We recall that the final prediction of \our{} is calculated by taking the average of logits returned by the target networks. Our analysis shows that non-informative augmentations tend to craft target networks that are "uncertain" of their predictions. In order not to influence the voting too much, their logits are centered around 0. In contrast, target networks generated by augmentations containing the informative feature return confident predictions. Since the pooling layer operates on logits, \our{} is able to reduce the effect of noisy models.  

\begin{figure}
    \centering
    \includegraphics[width=0.8\linewidth]{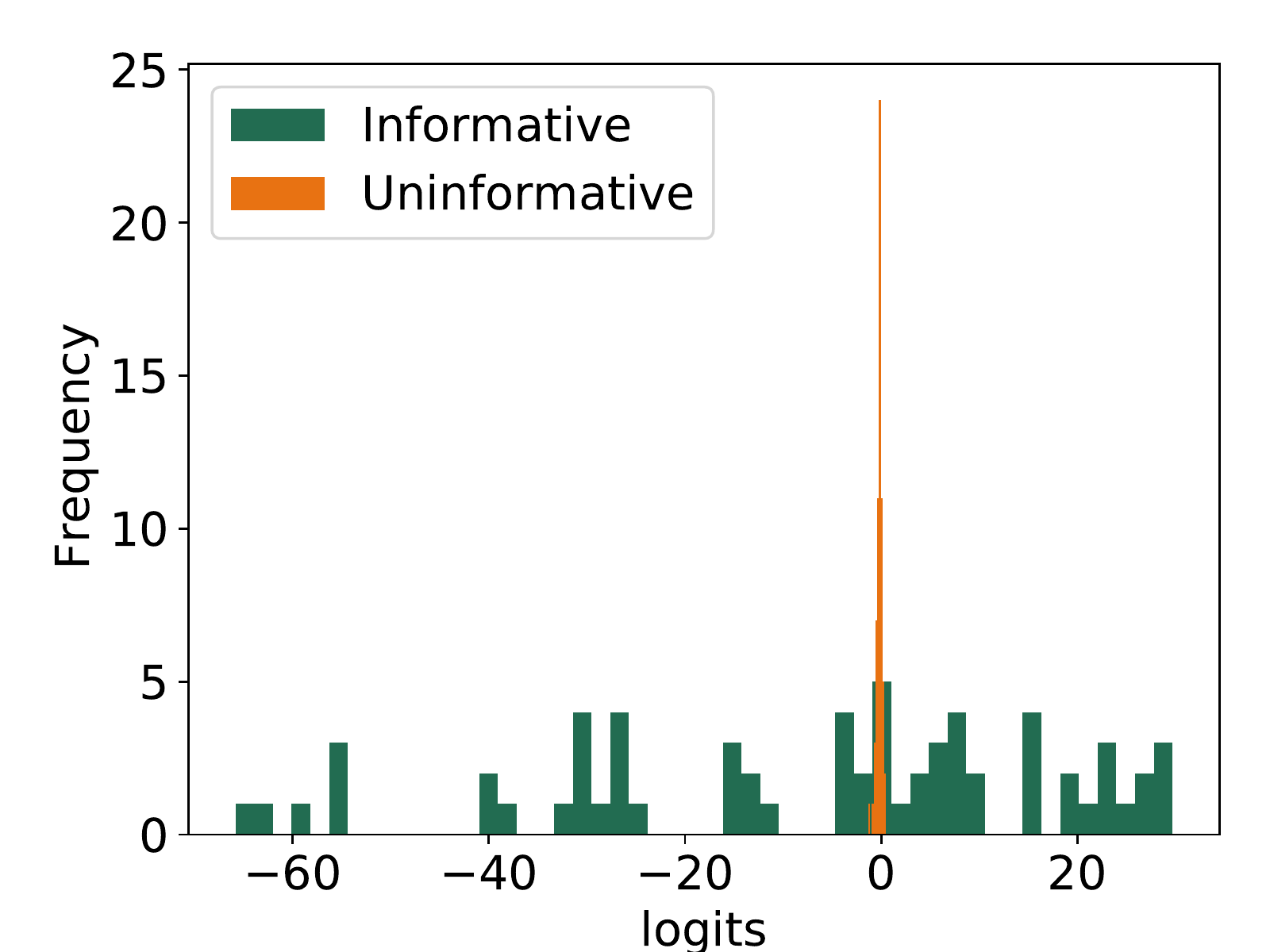}
    \caption{Histogram of target networks' predictions (logits) calculated on informative and non-informative augmentations. Since \our{} averages logits of target networks, the contribution of uncertain predictions generated by non-informative features is marginal compared to the confident scores obtained using informative features.}
    \label{fig:ablation}
\end{figure}


\section{Conclusion and future directions}

This paper introduced \our{} -- a hypernetwork-based ensemble for deep learning on tabular datasets. Making use of feature subsetting as data augmentations, we virtually increase the number of training examples, keeping the number of trainable weights unchanged. This is especially profitable for small datasets, where typical deep learning models perform inferior to shallow methods. Our experiments clearly confirm that \our{} obtains state-of-the-art results on small tabular datasets and performs on par with other methods on larger datasets. 

We strongly believe that this paper may serve as a stepping stone for hypernetwork-based approaches for tabular data, as there are still many venues that can be further pursued. \our{} relies on a random scheme for sampling augmentations. Although our analysis shows that \our{} is robust to irrelevant features, we can introduce a differentiable procedure to learn an informative and diverse set of augmentations (potential ideas can be found in \cite{node19}). In addition to the classification task, we plan to consider regression problems. Our analysis shows that the target networks tend to produce near-zero output if they are uncertain of the correct result, suggesting that \our{} may be suitable for regression.

\section*{Acknowledgement}

The research of M. \'Smieja was supported by the National Science Centre (Poland), grant no. 2022/45/B/ST6/01117. For the purpose of Open Access, the author has applied a CC-BY public copyright license to any Author Accepted Manuscript (AAM) version arising from this submission.

\appendix

\section{Datasets}

Overview of basic tabular datasets is presented in \Cref{table:dataset_description}. Real-life metagenomic datasets are described in \Cref{table:metagenomic_datasets}.

\begin{table}[t]
    \caption{Overview of the datasets. Here, $n, d, k$ denote the number of samples, features, and classes, respectively. Domains abbreviations: CM - customer metadata, TF - text features, RD - radar data, CD - clinical data, PX - pixels, Co - compositional, S - synthetic, VF - video features, IF - image features, VoF - voice features, Bio - biological dataset, AF - animal features}
    \label{table:dataset_description}
    \footnotesize
    \begin{tabular}{lccccc}
    \hline
     Dataset &  $n$ & $d$ &  Dom. & Source & $k$ \\
    \hline
     {Christine} & 5418 & 1637 & S & OpenML & 2 \\
     {CNAE-9} & 1080 & 857 & TM & UCI & 9 \\
     {Connectionist} & 208 & 60 & RD & UCI & 2 \\
     {Dermatology} & 366 & 33 & CD & UCI & 6 \\
     {Devanagari} & 12912 & 784 & PX & Kaggle & 58 \\
     {Fabert} & 8237 & 801 & S & OpenML & 7 \\
     {FashionMNIST} & 70000 & 784 & PX &  - & 10 \\
     {Glass} & 214 & 10 & Co & UCI & 7 \\
     {Heart Disease} & 303 & 14 & CD & UCI & 2\\
     {Hill-Valley} & 606 & 101 & S & UCI & 2 \\
     {Ionosphere} & 351 & 34 & RD & UCI & 2 \\
     {Libras} & 360 & 91 & VF & UCI & 15  \\
     {Lymphography} & 148 & 18 & IF & UCI & 2 
     \\
     {Mult. Features} & 2000 & 649 & IF & UCI & 10  \\
     {Nomao} & 34465 & 120 & CM & OpenML & 2 \\ 
     {OvarianTumour} & 283 & 54622 & Bio & OpenML & 3 \\
     {Parkinsons} & 197 & 23 & VF & UCI & 2 \\
     {Promoter} & 106 & 58 & Bio & UCI & 2 \\
     {Volkert} & 58 310 & 181 & S & OpenML & 10 \\
     {WBC} & 569 & 30 & IF & UCI & 2 \\ 
     {Zoo} & 101 & 17 & AF & UCI & 7 \\ 
    \hline
    \end{tabular}
\end{table}

\begin{table}[t]
    \centering
    \caption{Overview of the metagenomical datasets.}
    \label{table:metagenomic_datasets}
    \footnotesize
    \begin{tabular}{lcc}
    \hline
     Dataset &  $n$ & $d$ \\
    \hline
     {FengQ-2015} & 107 & 606 \\
     {GuptaA-2019} & 60 & 308 \\
     {HanniganGD-2017} & 55 & 292 \\
     {JieZ-2017} & 385 & 683 \\
     {KeohaneDM-2020} & 117 & 381 \\
     {LiJ-2017} & 155 & 436 \\
     {NagySzakalD-2017} & 100 & 438 \\
     {NielsenHB-2014} & 317 & 606 \\
     {QinJ-2012} & 344 & 651 \\
     {QinN-2014} & 237 & 645 \\
     {RubelMA-2020} & 175 & 370 \\
     {ThomasAM-2018a} & 53 & 477 \\
     {ThomasAM-2018b} & 60 & 503 \\
     {ThomasAM-2019c} & 80 & 519 \\
     {VogtmannE-2016} & 104 & 540 \\
     {WirbelJ-2018} & 125 & 537 \\
     {YachidaS-2019} & 509 & 718 \\
     {YuJ-2015} & 128 & 575 \\
     {ZellerG-2014} & 114 & 652 \\
     {ZhuF-2020} & 171 & 480 \\
    \hline
    \end{tabular}
\end{table}

\section{Hyperparameter optimisation}
\label{app:grid}
Each model's specific set of hyperparameters was evaluated on the test set. Models were optimized across the following hyperparameters: 

\subsection{XGBoost}
XGBoost was optimized with two grids, each containing a different set of parameters. The optimal hyperparameters obtained from the first grid were later used when performing a search on the second grid. \\
\newline
First grid:
    \begin{itemize}
        \item $n\_estimators$: \{50, 100, 250, 500, 1000, 3000\},
        \item $max\_depth$: \{2, 3, 5, 10, 15\},
        \item $learning\_rate$: Log-Uniform distribution [1e-5,1e-1],
        \item $min\_child\_weight$: \{1, 2, 4, 8, 16, 32\}
        \item $gamma$: \{0, 0.001, 0.1, 1\}.
    \end{itemize} \; \\    
Second grid:
    \begin{itemize}
        \item $subsample$: \{0.5, 0.6, 0.7, 0.8, 0.9, 1\},
        \item $reg\_lambda$: Log-Uniform distribution [1e-5, 10] initial\_value = 0,
        \item $reg\_alpha$: Log-Uniform distribution [1e-5, 10] initial\_value = 0.
    \end{itemize}
    
\subsection{NODE}
    \begin{itemize}
        \item $layer\_dim$: \{64, 128, 256, 512, 1024\}. \textit{In some cases value 1024 was omitted due to memory issues (big datasets)},
        \item $num\_layers$: Discrete uniform distribution [1, 5],
        \item $depth$: Discrete uniform distribution [2, 7],
    \end{itemize}

\subsection{DN}
    \begin{itemize}
        \item $epochs$: \{100, 150\},
        \item $dropout\_layer1$: \{0.1, 0.3, 0.5, 0.7\} (ordered),
        \item $dropout\_layer2$: \{0.1, 0.3, 0.5, 0.7\} (ordered),
        \item $dropout\_layer3$: \{0.1, 0.3, 0.5, 0.7\} (ordered),
        \item $dropout\_layer4$: \{0.1, 0.3, 0.5, 0.7\} (ordered),
        \item $learning\_rate$: \{3e-5, 3e-4, 3e-3, 3e-2, 3e-1\},
        \item $batch\_size$: \{32, 64\},
    \end{itemize}

\subsection{RF}
Common:
    \begin{itemize}
        \item $max\_features$: \{'sqrt', 0.2, 0.3, 0.5, 0.7\},
        \item $criterion$: \{'gini', 'entropy'\},
        \item $max\_depth$: \{'default', 2, 4, 8, 16\},
    \end{itemize}
Small datasets:
    \begin{itemize}
        \item $n\_estimators$: Discrete uniform distribution \{50, 3000\}, quantized to increments of 50,
    \end{itemize}
Big datasets:
    \begin{itemize}
        \item $n\_estimators$: \{ 50, 100, 200, 500, 1000, 3000 \}
    \end{itemize}

\subsection{HyperTab}
Common:
    \begin{itemize}
        \item $epochs$: \{100\},
        \item $target\_size$: \{5, 10, 20, 50\},
        \item $learning\_rate$: \{3e-5, 3e-4, 3e-3, 3e-2, 3e-1\}, 
    \end{itemize}
Big datasets:
    \begin{itemize}
        \item $masks\_no$: \{3, 5, 7, 20\},
        \item $mask\_size$: \{30\%, 50\%, 70\%, 80\%\} of \textit{n\_features},
    \end{itemize}
In the case of Small datasets, it was dependent on the dataset itself. Here we provide a generalized grid:
    \begin{itemize}
        \item $masks\_no$: Discrete uniform distribution [10, 200], quantized to increments of 10,
        \item $mask\_size$: Discrete uniform distribution [2, \textit{n\_features} * 0.9]
    \end{itemize}

\bibliographystyle{IEEEtran}
\bibliography{main}

\end{document}